\documentclass[11pt]{article}
\usepackage[preprint]{acl}
\usepackage{times,latexsym}
\usepackage[T1]{fontenc}
\usepackage[utf8]{inputenc}
\usepackage{microtype,amsmath,amssymb}
\usepackage{graphicx,booktabs,tabularx,array}
\usepackage{tikz,pgfplots}
\usetikzlibrary{positioning,arrows.meta,calc}
\pgfplotsset{compat=1.18}
\newcolumntype{P}[1]{>{\raggedright\arraybackslash}p{#1}}
\newcolumntype{Y}{>{\raggedright\arraybackslash}X}
\newcommand{\R}{\mathbb{R}}
\definecolor{rep}{HTML}{246A86}
\definecolor{mech}{HTML}{755393}
\definecolor{diag}{HTML}{39764E}
\definecolor{ctrl}{HTML}{B36128}
\title{Sparse Autoencoders for Mechanistic Interpretability in NLP:\\
A Critical Survey of Recoverability, Interpretability, and Utility}
\author{
    Salima Lamsiyah \qquad Adelaide Danilov \\
    \bf Oleksandr Marchenko Breneur \qquad Aria Nourbakhsh \\
    Department of Computer Science (DCS), University of Luxembourg \\
    \small{\textbf{Correspondence:} \href{mailto:salima.lamsiyah@uni.lu}{salima.lamsiyah@uni.lu}} \\
}

\newcounter{evidencecase}
\renewcommand{\theevidencecase}{E\ifnum\value{evidencecase}<10 0\fi\arabic{evidencecase}}
\newcommand{\reviewmaterials}{\href{https://anonymous.4open.science/r/sae-research-register-C331/}{anonymous research register}}

\hypersetup{
  colorlinks=true,
  citecolor=blue,
  linkcolor=blue,
  urlcolor=blue,
  pdfauthor={Anonymous},
  pdftitle={Sparse Autoencoders for Mechanistic Interpretability in NLP: A Critical Survey of Recoverability, Interpretability, and Utility}
}

\begin{document}

\maketitle

\begin{abstract}
Sparse autoencoders (SAEs) have emerged as a promising
approach to mechanistic interpretability by decomposing
language-model activations into sparse, potentially
interpretable features. However, differences in design
assumptions, evaluation objectives, and experimental
settings complicate conclusions about their effectiveness.
We present a critical survey of 138 studies on SAEs in
natural language processing. Our contributions are fourfold:
(1) we propose a task-grounded taxonomy connecting dictionary
design, feature operations, and NLP functions;
(2) we examine how sparsity allocation, geometry, data
coverage, and inference affect feature recoverability;
(3) we assess evidence for semantic interpretability,
mechanistic faithfulness, and behavioral utility; and
(4) we synthesize findings across multilingual processing,
retrieval, reasoning, classification, and behavioral
regulation, highlighting the roles of objectives, baselines,
supervision, and intervention pipelines.
Through comparative analyses of published results,
we reconcile apparent disagreements, identify unresolved
questions, and propose directions for more rigorous
evaluation. An anonymous research
register\footnote{\url{https://anonymous.4open.science/r/sae-research-register-C331/register.html}}
accompanies the survey.
\end{abstract}

\section{Introduction}
Sparse autoencoders encode model activations as sparse coefficients over a learned dictionary of feature directions \citep{O086}. They make it possible to inspect concepts, compare representations, and intervene on internal variables at a finer granularity than whole layers or attention heads. Sparse interpretable sentence embeddings provide an earlier precedent \citep{O091}; transformer studies established useful feature decompositions \citep{A03,O086}; subsequent work scaled dictionaries and released reusable resources \citep{A04,O081,O098}. The scientific opportunity is now broader: can these coordinates explain how language is represented and processed, and can that understanding improve NLP systems?

Answering this question requires separating four outcomes. \emph{Recoverability} concerns whether an encoding procedure recovers the target latent structure, where that structure is specified. \emph{Semantic interpretability} concerns whether a feature description makes discriminating predictions about activations or effects. \emph{Mechanistic faithfulness} concerns whether an explanation identifies computations the original model uses. \emph{Behavioral utility} concerns success on a stated task with its competing objectives. These outcomes can agree, but their definitions do not imply agreement.

Consider language identity. Multilingual feature analyses use it to distinguish language-associated representations \citep{O007}; language steering seeks to induce a chosen output language \citep{N67}; LangSAE removes it from pooled embeddings to improve multilingual retrieval \citep{N64}. Grouping these studies simply as multilingual SAE applications hides an objective-dependent reversal: information useful for generation can be a nuisance for semantic retrieval. Conversely, grouping only by encoder architecture does not reveal this distinction at all.

Existing surveys provide complementary perspectives on SAEs and mechanistic interpretability. \citet{O027} synthesize SAE architectures, training procedures, feature explanations, evaluation methods, and applications. \citet{A06} organize NLP mechanistic interpretability around causal mediators and search methods, examining which computational units and analytical procedures support explanation, hypothesis testing, and model editing. \citet{O078} connect localization and steering to improvements in alignment, capability, and efficiency, including SAE features, task-specific metrics, and checks for unintended effects. Building on these foundations, our survey examines how SAE construction and use jointly shape the comparability and interpretation of NLP findings.

This motivates three research questions. \textbf{RQ1: Which design assumptions change what an SAE can recover?} We compare allocation, geometry, data coverage, and inference under their stated objectives and theoretical assumptions (Section~\ref{sec:design}). \textbf{RQ2: What evidence distinguishes an interpretable or useful feature from a faithful mechanism?} We examine semantic tests, intervention controls, and counterexamples using a claim-specific rubric (Section~\ref{sec:meaning}). \textbf{RQ3: Why do NLP studies reach different conclusions about SAE usefulness?} We compare task objectives, supervision, and control pipelines, and reanalyse selected within-study results (Section~\ref{sec:nlp}).

Our taxonomy organizes SAE studies by dictionary design, feature
operation, and NLP function. Our multi-label register and comparative analyses support a \textbf{task-grounded synthesis} that clarifies apparent disagreements and identifies unresolved questions.

\section{Scope and Organizing Framework}
\label{sec:map}

We survey SAE methods for language-model analysis
and NLP system construction, including relevant speech and
multimodal extensions. Corpus selection and coding are detailed
in Appendix~\ref{app:method}.

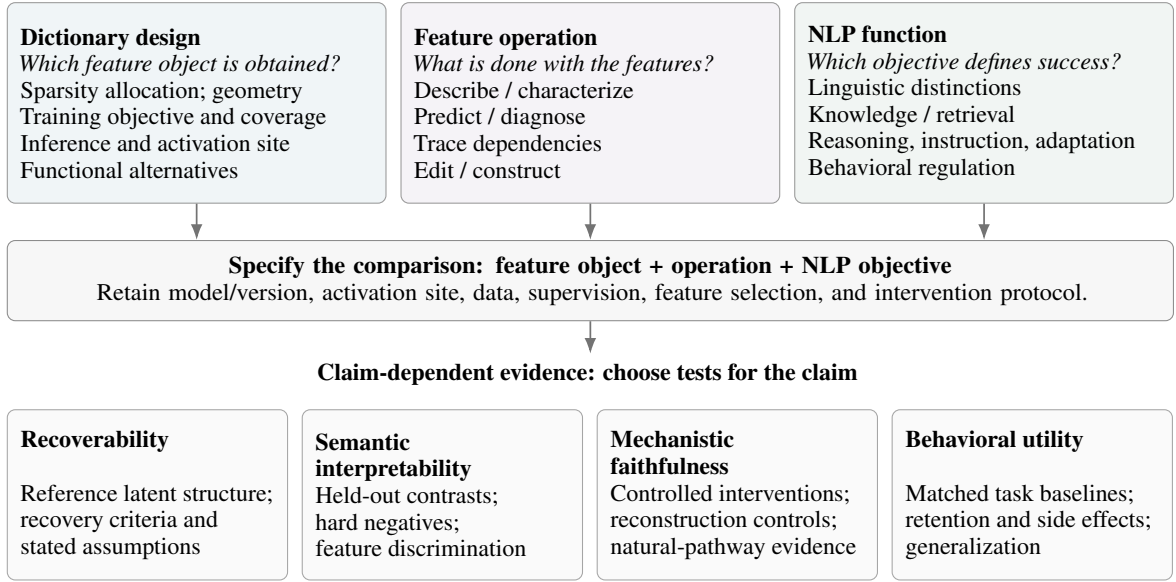
\begin{figure*}[t]

\centering
\begin{tikzpicture}[
  font=\small,
  >=Latex,
  facet/.style={
    draw=black!40,
    rounded corners=3pt,
    align=flush left,
    inner sep=5pt,
    text width=4.65cm,
    minimum height=2.65cm,
    anchor=north west
  },
  claim/.style={
    draw=black!35,
    rounded corners=3pt,
    align=flush left,
    inner sep=5pt,
    font=\footnotesize,
    text width=3.35cm,
    minimum height=2.25cm,
    anchor=north west
  },
  flow/.style={
    ->,
    draw=black!55,
    line width=.65pt
  }
]

% Parallel facets, rather than sequential research stages.
\node[facet,fill=rep!7] (design) at (0,0) {
  \textbf{Dictionary design}\\
  \emph{Which feature object is obtained?}\\
  {\footnotesize
  Sparsity allocation; geometry\\
  Training objective and coverage\\
  Inference and activation site\\
  Functional alternatives}
};

\node[facet,fill=mech!7] (operation) at (5.2,0) {
  \textbf{Feature operation}\\
  \emph{What is done with the features?}\\
  {\footnotesize
  Describe / characterize\\
  Predict / diagnose\\
  Trace dependencies\\
  Edit / construct}
};

\node[facet,fill=diag!7] (function) at (10.4,0) {
  \textbf{NLP function}\\
  \emph{Which objective defines success?}\\
  {\footnotesize
  Linguistic distinctions\\
  Knowledge / retrieval\\
  Reasoning, instruction, adaptation\\
  Behavioral regulation}
};

% The three facets jointly specify the comparison.
\node[
  draw=black!45,
  rounded corners=3pt,
  fill=black!3,
  align=center,
  inner sep=6pt,
  text width=14.98cm,
  below=.48cm of operation
] (compare) {
  \textbf{Specify the comparison:
  feature object + operation + NLP objective}\\
  {\footnotesize
  Retain model/version, activation site, data, supervision,
  feature selection, and intervention protocol.}
};

\draw[flow]
  (design.south) -- (design.south |- compare.north);
\draw[flow]
  (operation.south) -- (compare.north);
\draw[flow]
  (function.south) -- (function.south |- compare.north);

\node[
  below=.43cm of compare,
  align=center
] (evidence) {
  \textbf{Claim-dependent evidence:
  choose tests for the claim}
};

\draw[flow] (compare.south) -- (evidence.north);

% Complementary assessment targets, with no ranking.
\node[claim,fill=black!2] (recovery)
  at ($(design.west |- evidence.south)+(0,-.18)$) {
  \textbf{Recoverability}\\
  \strut\\
  Reference latent structure;\\
  recovery criteria and\\
  stated assumptions
};

\node[
  claim,
  fill=black!2,
  right=.18cm of recovery.north east,
  anchor=north west
] (meaning) {
  \textbf{Semantic}\\
  \textbf{interpretability}\\
  Held-out contrasts;\\
  hard negatives;\\
  feature discrimination
};

\node[
  claim,
  fill=black!2,
  right=.18cm of meaning.north east,
  anchor=north west
] (mechanism) {
  \textbf{Mechanistic}\\
  \textbf{faithfulness}\\
  Controlled interventions;\\
  reconstruction controls;\\
  natural-pathway evidence
};

\node[
  claim,
  fill=black!2,
  right=.18cm of mechanism.north east,
  anchor=north west
] (utility) {
  \textbf{Behavioral utility}\\
  \strut\\
  Matched task baselines;\\
  retention and side effects;\\
  generalization
};

\end{tikzpicture}
\caption{Three facets jointly define a comparison (top), interpreted
alongside its experimental settings (middle). Complementary evidence
targets assess what the findings support (bottom). Arrows show how
the facets inform study comparison and claim assessment; the tests
are illustrative and tailored to the claim.}
\label{fig:taxonomy}
\end{figure*}

\paragraph{Three facets and their qualifiers.}
For study $s$, we record $T(s)=(D_s,O_s,F_s;Z_s)$
(Figure~\ref{fig:taxonomy}). \textbf{Dictionary design} $D_s$
describes how features are obtained or constrained: sparsity
allocation, geometry, coverage and objectives, inference and
activation sites, or functional alternatives.
\textbf{Feature operation} $O_s$ describes their use: characterize
representations, predict or diagnose outcomes, trace dependencies,
or edit model behavior and construct systems. 
NLP function $F_s$ specifies the evaluation target,
covering linguistic distinctions, knowledge and retrieval,
reasoning and adaptation, behavioral regulation,
text analytics, speech and multimodal analysis,
and general enabling research.

The facets interact: a task can
shape dictionary training, while dictionary structure can enable
particular operations. Qualifiers $Z_s$ retain model/version,
activation site, data, supervision, feature selection, and
intervention protocol. Together, the upper boxes and central band
in Figure~\ref{fig:taxonomy} expose differences in experimental
settings that may explain apparent disagreements.

\paragraph{Claim-dependent evidence.}
The taxonomy identifies comparable studies; the evidence framework assesses their claims. In the lower row of Figure~\ref{fig:taxonomy}, claims about feature
recoverability, semantic interpretability,
mechanistic faithfulness, and behavioral utility
require different tests and controls. In particular,
demonstrating the utility of SAE-derived features in
a classifier, retriever, or controller does not
necessarily establish the computations performed
by the original model.

The operational definitions, study assignments,
and cross-dimensional analyses are provided in
Appendix~\ref{app:taxonomy}.

\section{Dictionary Design and Recoverability}
\label{sec:design}

RQ1 asks which design assumptions determine what an SAE can recover.
For an NLP researcher seeking a grammatical distinction, an
unsuccessful feature search leaves several explanations open:
the chosen representation, the dictionary's constraints, the training
data, or the encoding procedure. We compare methods by which of
these limitations they address.

For activation $h\in\R^d$, an encoder produces
$z=f_\theta(h)\in\R^m$ and a decoder reconstructs $\hat h=b+Dz$.
A common objective is
\begin{equation}
\mathcal L=
\mathbb E[\|h-b-Df_\theta(h)\|_2^2+\lambda\Omega(z)].
\label{eq:sae}
\end{equation}
This objective rewards reconstruction and sparsity; recovering a
specified latent structure is a separate endpoint. Synthetic
studies can compare against a known generating dictionary, whereas
language-model studies must validate the proposed feature
interpretation.

\subsection{Defining the structure to be recovered}

The activation site determines the starting object. Residual streams
contain accumulated information; attention and MLP outputs
characterize component contributions. Transcoders instead
approximate a component's input--output map \citep{O050,O100}.
Layer transfer, shared dictionaries, and routing change training
cost and parameter reuse \citep{O003,O016,O015}; comparisons still
need to establish what structure is preserved across sites.

Within a site, the question is how that structure should be
represented. Splitting distributes a concept across latents, while
absorption allows a specialized latent to carry a broader property
\citep{N03}; neither supports equating an inactive coordinate with
absent information. Subspace \citep{O049,N20} and manifold
approaches \citep{O109} model structure beyond individual
directions. Nested scales \citep{O114}, parent--child gates
\citep{N13}, and Kronecker factorization \citep{N73} impose
different forms of organization across features. Temporal
smoothness \citep{O069} and predictable-context/novelty separation
\citep{N27} instead organize variation over sequences. These
choices change the candidate unit of analysis: their NLP value
depends on whether the resulting directions, groups, or temporal
features capture the distinction under study.

\subsection{Allocating sparse capacity}

Once the target is specified, allocation determines which features
can participate in each reconstruction. An $\ell_1$ penalty
encourages sparsity but also shrinks amplitudes. Gated SAEs
separate selection from magnitude estimation, while JumpReLU
learns activation thresholds \citep{O087,O053}. TopK fixes the
number of retained coordinates per input \citep{O081}; BatchTopK
distributes a budget across a batch and estimates an inference-time
threshold \citep{A05}. Top-AFA selects counts through a
quasi-orthogonality approximation \citep{A02}, AdaptiveK uses a
learned complexity signal \citep{N58}, and Group Bias Adaptation
(GBA) targets activation-frequency coverage \citep{N40}. These
allocation rules make different assumptions about inputs,
geometry, and feature frequency.

The results show why allocation cannot be ranked by reconstruction
alone. BatchTopK improves reconstruction over TopK, but its
Gemma-2-2B comparison with JumpReLU is mixed for downstream
cross-entropy degradation \citep{A05}. In GBA's variable-sparsity
synthetic experiment, TopK recovery falls from 98.4\% at $K=20$
to 23.4\% at $K=50$; full versus misspecified frequency coverage
gives GBA 100\% versus 3.9\% recovery \citep{N40}. These rates use
alignment threshold 0.9 and 128 generating features. They
demonstrate sensitivity to allocation and coverage assumptions
under known ground truth, rather than an NLP ranking.
Comparisons should match average sparsity, data, and insertion
protocol; decoder normalization must also be recorded because it
changes coefficient scales (Appendix~\ref{app:technical}).

\subsection{Separating coverage from inference error}

Even a suitable representation and budget cannot resolve every
failure. Specialized sampling and tilted risk improve access to
rare concepts \citep{O031}, changing the training distribution and
emphasis of the objective. Weighted SAEs analyze recovery and
reweighting under an idealized tied-weight model \citep{N30};
those assumptions delimit the result's applicability to
language-model activations. Locally amortized inference addresses
a different limitation: a shared encoder can leave an optimization
gap relative to per-input sparse coding \citep{N35}. Improving
that approximation does not establish that the reference codes
are the intended linguistic features. Thus data coverage, the
learning objective, and encoding accuracy require separate
comparisons.

\paragraph{Answer to RQ1.}
Recoverability has separable bottlenecks: the representation and
its geometry, sparse allocation, training coverage, and inference.
The evidence supports diagnosing the suspected bottleneck while
matching the remaining conditions. Reconstruction, recovery
against a specified reference, and behavioral preservation should
remain separate endpoints. Technical comparisons and assumptions are detailed in Appendix~\ref{app:technical}.

\section{From Features to Explanations and Mechanisms}
\label{sec:meaning}

\begin{table*}[t]
\centering
\footnotesize
\setlength{\tabcolsep}{4pt}
\renewcommand{\arraystretch}{1.18}

\begin{tabularx}{\textwidth}{
@{}
>{\raggedright\arraybackslash}p{0.17\textwidth}
>{\raggedright\arraybackslash}p{0.20\textwidth}
>{\raggedright\arraybackslash}p{0.24\textwidth}
>{\raggedright\arraybackslash}X
@{}
}
\toprule
\textbf{Claim} &
\textbf{Weak evidence} &
\textbf{Moderate evidence} &
\textbf{Stronger evidence} \\
\midrule

Feature description
&
A label supported only by selected activating examples.
&
The description predicts feature activation on unseen
positive and negative examples.
&
Tests with hard negatives and counterexamples;
the description distinguishes the feature from
similar features.
\\

\addlinespace[3pt]
Prediction / diagnosis
&
Correlation observed on training data.
&
Predictive performance on held-out data,
compared with a relevant baseline.
&
Generalization and calibration tests;
matched supervision and checks for
confounding or data leakage.
\\

\addlinespace[3pt]
Mechanistic role
&
Feature coactivation or behavioral changes
without adequate controls.
&
A targeted intervention produces the
predicted behavioral effect.
&
Semantic counterfactuals, reconstruction-error
controls, and tests of alternative computational
pathways or circuit dependencies.
\\

\addlinespace[3pt]
Task control
&
A few selected examples of successful
behavior modification.
&
Task improvement over a baseline,
with effects on other capabilities assessed.
&
Matched feature-selection and control
baselines, with robustness, side effects,
and persistence evaluated.
\\

\bottomrule
\end{tabularx}

\caption{Evidence supporting different claims about SAE
features. Stronger evidence requires more discriminating
tests and appropriate controls. The criteria apply to
specific claims under stated evaluation conditions,
not to the overall quality of a paper.}
\label{tab:rubric}
\end{table*}

Recovering sparse features leaves their explanatory role unresolved. RQ2 asks what distinguishes a meaningful or useful feature from a faithful account of model computation. A feature may predict sentiment, support sentiment steering, or mediate the model's own sentiment decision; these claims require different tests. Table~\ref{tab:rubric} organizes support for a specified claim and evaluation setting. Appendix~\ref{app:evidence} develops the assessment rules through worked claims and a causal counterexample.

\subsection{Testing the feature's meaning}

A feature description is a hypothesis about its behavior. Input-oriented descriptions predict what elicits activation; output-oriented descriptions predict contributions to generation \citep{O095}. Automated explanation \citep{O097}, agentic testing \citep{O036}, and counterexample-guided revision \citep{O089} make these hypotheses testable on held-out positives and hard negatives. The critical question is whether the description discriminates the intended property from alternatives. Descriptive collision tests whether one explanation also fits unrelated features \citep{O111}; contrastive stories, semantic/functional neighborhoods, and concept-separability tests examine complementary aspects of specificity \citep{O012,O014,O040,O088}.

The controls must preserve what the claim says should remain invariant. A grammatical feature should track the specified contrast across appropriate lexical controls; a lexical feature need not survive word substitution. Adversarial inputs test the limits of a description \citep{O037}, while randomized SAE components and randomized host models challenge different sources of apparent interpretability \citep{N19,N28}. In the latter comparison, similar aggregate scores in several tested settings show that those metrics need not distinguish learned computations from structure present without training \citep{N28}. Cross-run consistency asks whether features recur across learned dictionaries \citep{N46}; recurrence alone does not validate their meaning.

\subsection{Testing predictive and causal effects}

A useful predictor need not identify a variable the source model uses. For a known target, SAE probes should be compared with supervised probes on the original activations, given matched information \citep{O063,O084}. For control, direct activation directions provide an alternative \citep{N67}. These comparisons test the benefit of the SAE representation for the chosen objective. A reusable feature vocabulary can still support exploratory discovery beyond a predefined prediction task \citep{O101}.

Interventions test what changes when a feature is edited. Their effects must be separated from damage introduced by reconstruction. With error $e=h-(b+Dz)$, an error-preserving edit is
\begin{equation}
h'=b+Dz'+e=h+D(z'-z).
\label{eq:edit}
\end{equation}
Replacing $h$ with $b+Dz'$ also removes $e$, so the measured effect combines the feature edit with reconstruction error. Retaining $e$ isolates the decoded change but does not ensure that the edited activation lies within the model's natural distribution. Null edits, matched-magnitude controls, naturally occurring replacement values, and off-target measurements test distinct alternatives. A controlled effect supports a causal claim about the manipulation; explaining the unmodified computation requires further evidence.

\subsection{Testing the proposed mechanism}
The gap is explicit in \citet{A01}. Their patching direction combines a concept-correlated component that is disconnected from the output with an ordinarily dormant, output-sensitive component. Patching transfers a signal between them and changes the answer through a route unused before intervention. The successful edit therefore fails to identify the natural mechanism; Appendix~\ref{app:illusion} instantiates this counterexample. Conversely, redundant pathways can make a genuinely used feature dispensable under ablation. Necessity and sufficiency must consequently be interpreted relative to the tested interventions and contexts.

A circuit claim further specifies dependencies between features. Sparse feature circuits retain feature and error nodes and test task-relevant effects \citep{O083}. Module-level interventions on coactivation groups \citep{N47} and transport-based feature matching \citep{N21} concern group effects and correspondence. Establishing a directed circuit also requires tests of the proposed downstream dependencies; edge attribution helps select candidates for those tests. Attention-output features, formal languages, and Othello models provide distinct computational testbeds whose scope must accompany the interpretation \citep{O051,O096,O046}. Neurons, heads, key--value features, and functional dictionaries offer alternative units for the computational account \citep{O048,O105,N06,N52,O050}; the comparison should establish what the SAE basis adds.

\paragraph{Answer to RQ2.}
Semantic interpretability rests on discriminating predictions, and utility on appropriate task comparisons. Mechanistic faithfulness requires evidence that the proposed variables and dependencies capture computation the original model uses. Controlled edits contribute to that assessment, but their success alone does not establish the claimed pathway. A feature can remain useful for control even when its mechanistic interpretation fails.

\section{Comparing SAE Usefulness Across NLP Tasks}
\label{sec:nlp}
RQ3 asks why NLP studies reach different conclusions about SAE usefulness.  The comparison depends on what the task requires, what information the method receives, and which alternative it must outperform. We organize the findings around these differences to identify both useful contributions and the conditions under which their advantages hold. Appendix~\ref{app:cases} provides extended comparisons.

\subsection{Objectives change the value of a feature}
Language identity makes the role of the objective explicit. LangSAE removes language-associated components from pooled multilingual-e5 embeddings for semantic retrieval, whereas CLaS-Bench edits Llama-3.1-8B-Instruct activations to control the output language \citep{N64,N67}. Suppressing language identity and strengthening it serve different goals. Their findings concern different task requirements.

Within CLaS-Bench, residual difference-in-means scores 84.5 versus 42.3 for SAE difference-in-means, a 42.2-point gap in the combined language/relevance metric \citep{N67}. The reported selected settings use layers 20 and 25, respectively, both at strength 5. This establishes an advantage for the selected residual-space pipeline; differences in available layers and selection prevent isolating the SAE basis as the cause (Appendix~\ref{app:quant}).

Linguistic analysis asks a more specific question than language control. Language-associated groups support observation and intervention \citep{O007}; LinguaLens tests Chinese/English linguistic contrasts \citep{O019}, while TenseLoC localizes tense and tests English-feature steering \citep{O032}. Script-sensitive representations motivate matched-content and transliteration controls, with raw-neuron interventions serving as a direct comparator \citep{N51}. Italian-model analysis extends coverage \citep{O013}. Together, these studies motivate testing whether corresponding features preserve a specified linguistic contrast and intervention effect across languages; broad language selectivity or geometric similarity alone does not establish that transfer.

\subsection{Representations and baselines define the comparison}
Classification provides a concrete reconciliation. Positive Gemma results compare SAE classifiers with last-token hidden-state baselines \citep{O021}. In sparse probing, adding attention-pooled dense baselines reduces the apparent advantage of pooled SAE probes \citep{O063}. Both studies examine pooling, so its presence alone cannot explain their disagreement. The within-study comparison establishes that baseline construction matters; attributing all cross-paper differences to it would require matching the model, data, feature selection, and classifier. Task-aligned classification additionally changes representation training \citep{O066}.

DILA and topic modeling construct interpretable task systems \citep{O056,N68}, rather than explaining an unchanged source model. Speech analysis, vision--language alignment, and multimodal tracing further change the representation being examined \citep{O110,N18,N34,N36,O107}. These extensions require comparisons appropriate to acoustic frames, visual embeddings, or text tokens.

\subsection{Information access separates RAG outcomes}
Retrieval-oriented objectives can preserve useful matching structure \citep{O034}, while adapting a retriever around a frozen dictionary constructs a new retrieval system \citep{N25}. For retrieval-augmented generation (RAG), improvements concern different stages: detecting unsupported output, choosing between context and memory, enriching a query, or repairing an answer. Table~\ref{tab:rag} distinguishes these operations and their information requirements. A detector can identify errors without explaining their cause, and increased reliance on context helps only when that context is trustworthy.

\begin{table*}[t]
\centering
\footnotesize
\setlength{\tabcolsep}{4pt}
\renewcommand{\arraystretch}{1.15}

\begin{tabularx}{\textwidth}{
@{}
>{\raggedright\arraybackslash}p{0.19\textwidth}
>{\raggedright\arraybackslash}p{0.24\textwidth}
>{\raggedright\arraybackslash}p{0.22\textwidth}
>{\raggedright\arraybackslash}X
@{}
}
\toprule
\textbf{SAE use} &
\textbf{Information used} &
\textbf{Reported contribution} &
\textbf{Key evaluation distinction} \\
\midrule

Hallucination detection and feedback
\citep{O090,N32}
&
Sparse features and detection signals;
feedback adds a generation stage.
&
Detect unsupported answers and guide
subsequent generation.
&
Evaluate detection and generation
improvements separately.
\\

\addlinespace[3pt]
Context--memory selection
\citep{O006,O033}
&
Features identified through contrasts
between retrieved context and
internal knowledge.
&
Change which information source
the model relies on.
&
Measure source preference and
answer correctness separately.
\\

\addlinespace[3pt]
Query enrichment
\citep{O029,N80}
&
SAE-guided query modifications;
some analyses select successful
query expansions.
&
Improve retrieval/generation or
analyze query-expansion effects.
&
Test on unselected queries and
evaluate the contribution of the
SAE within the full pipeline.
\\

\addlinespace[3pt]
Distractor repair
\citep{N55}
&
Known correct answer token used
to optimize latent activations on
selected difficult examples.
&
Recover answers affected by
distracting documents.
&
Distinguish oracle-assisted repair
from autonomous error correction.
\\

\bottomrule
\end{tabularx}

\caption{Comparison of SAE-based approaches to
RAG. The methods differ
in their objectives, available information, and
evaluation settings; their reported improvements
therefore do not measure the same capability.}
\label{tab:rag}
\end{table*}

LatDefuse illustrates why information access belongs in the comparison. It optimizes latent edits toward a known correct token at answer divergence and evaluates distractor-vulnerable instances \citep{N55}. The result establishes conditional repair capacity. Comparing its recovery rate directly with autonomous detection or correction would conflate different supervision and evaluation populations. An integrated system would need to evaluate detection, source arbitration, and repair together, including errors passed between stages.

\subsection{Selection and prompts shape instruction gains}
For reasoning, steering a strategy and identifying its computation are separate achievements. Reasoning-associated features can change traces or strategies \citep{O076,O104}, while cue-injection and meaning-preserving tests can challenge an abstract-reasoning label \citep{O072}. Temporal-difference analysis ties features to specified computational variables in an in-context reinforcement-learning task \citep{O082}. These findings answer different questions about control, feature meaning, and computation; lexical cues must be varied independently of the required operation to compare those claims.

Instruction methods also differ in how they select features: SAIF identifies instruction-related features \citep{O062}, narrative editing generates compliant/violating contrasts \citep{O017}, SISTER uses prompt variation \citep{O020}, and PULSE uses labeled demonstration utility \citep{N17}. Comparators need access to the same signals. Evaluation prompts matter as well: for Gemma-2-2B sentiment circuits, SST-2 F1 gains are 37.19 points under template T1 and 2.96 under T4 \citep{N49}. These three-run mean differences establish prompt dependence within the reported setting; they do not isolate baseline headroom as its cause.

The stage of use changes the comparison further. Concept-based pretraining \citep{N23} and feature-guided fine-tuning data selection \citep{N61} construct changed models; transfer prediction \citep{N41} and sparse model diffing \citep{O022} respectively anticipate and diagnose updates. Inference-time activation or input edits leave weights fixed; diffusion-language analyses additionally depend on denoising time and mask position \citep{O106}.

\subsection{Controllers determine regulation trade-offs}
Behavioral regulation tests what the complete intervention achieves. Refusal-feature analysis and adversarial detection \citep{O028} concern different outcomes from maintaining benign capabilities during control. Static refusal amplification can impair unrelated capabilities \citep{O060}; REINS and graph-regularized steering combine selected directions with runtime gating \citep{N71,N12}. Their comparison therefore involves both the feature directions and the decision about when to intervene.

Unlearning makes this distinction concrete. Static suppression has mixed multi-topic results \citep{O059}; Dynamic SAE Guardrails add input-dependent classification \citep{O092}, whereas subspace-guided projection and CRISP modify parameters \citep{O018,N59}. Conditional intervention can avoid unnecessary edits, while parameter updates remove dependence on a runtime controller. Comparing their forget--retain trade-offs requires matching target knowledge, evaluation populations, retention tasks, and relearning attacks. Adaptation constraints \citep{N60,N66} and backdoor analyses \citep{N77,N82} add further training and attack conditions. These differences identify what a controlled comparison must hold fixed before attributing a gain to the dictionary.

\paragraph{Answer to RQ3.}
SAE usefulness depends on the objective, representation, comparison baseline, available supervision, and intervention pipeline. Opposing treatments of language identity can both serve their respective tasks; classification, repair, and regulation results require matching the conditions defining success. The synthesis reconciles the scope of these findings. Establishing which factor causes a remaining performance difference requires experiments that vary it while controlling the others.

\section{Research Directions and Practical Choices}
\label{sec:synthesis}
The answers to RQ1--RQ3 identify recovery bottlenecks, requirements for mechanistic claims, and conditions shaping task utility. They motivate experiments that distinguish competing explanations and guide the choice of an appropriate unit of analysis.

\subsection{Three testable comparisons}
\paragraph{Discourse structure versus temporal persistence.}
Compare static and temporal dictionaries with matched host model, layer, capacity, average sparsity, and training corpus \citep{O069,N27}. Vary semantic continuity, lexical repetition, and abrupt discourse change independently. Measure held-out contrast discrimination and predicted intervention effects on entity/state tracking. This tests whether temporal structure captures discourse-relevant distinctions beyond lexical persistence; smooth codes and separate task gains would leave that relationship unresolved.

\paragraph{Feature matching versus intervention transfer.}
Match features across random seeds, paraphrases, and languages using geometry and activation distributions, then test whether correspondence predicts preservation of an independently specified intervention effect. Include unmatched and frequency-matched controls. Current consistency and matching measures \citep{N46,N21} motivate these comparisons, while known-concept disentanglement results \citep{N48} motivate testing functional roles separately. 

\paragraph{Selection and controller versus dictionary.}
With fixed train/validation/test splits, cross SAE dictionaries with shared supervised or unsupervised selection rules and static or gated controllers. Include dense probes for prediction and direct-space directions for control, with matched labels and tuning budgets. Report the target--retention frontier. The probing/classification disagreement and conditional-control advances motivate this design; attributing their differences to selection or gating remains a hypothesis until those factors are controlled.

\subsection{Choosing when an SAE adds value}
For a known predictive target, begin with a well-tuned supervised probe and a matched label budget. For output control, include direct activation directions before incurring dictionary-training and feature-selection costs. For a localized computational hypothesis, neurons, heads, or a functional surrogate may provide the required unit. Direct comparators support these choices \citep{N06,O105,N52,N67}. An SAE is more compelling when a reusable, fine-grained feature vocabulary enables discovery, comparison across many questions, or a demonstrated specificity/cost advantage. 

\section{Conclusion}
This survey critically examines the role of sparse
autoencoders in mechanistic interpretability and NLP.
Through a task-grounded taxonomy of dictionary design,
feature operations, and NLP functions, we identify
key challenges in feature recoverability, distinguish
semantic interpretability from mechanistic faithfulness,
and explain how task objectives and experimental
pipelines shape reported utility. Our comparative
synthesis reconciles apparent disagreements across
studies and highlights where matched evaluations
remain necessary. These findings provide NLP
researchers with a framework for evaluating SAE
contributions, selecting appropriate baselines,
and determining whether useful features also
explain the model's underlying computations.

\section*{Limitations}
This survey focuses on sparse autoencoders for
mechanistic interpretability and their applications
to natural language processing. It does not aim
to comprehensively cover related developments
in sparse coding, disentangled representation
learning, or interpretability methods beyond SAEs.
Given the rapid evolution of this research area,
our synthesis reflects the literature available
within the survey's defined scope.
Comparisons across studies are constrained by
differences in model architectures, datasets,
feature representations, and evaluation protocols.
Accordingly, our analyses emphasize the conditions
under which findings are supported rather than
establishing universal rankings of SAE methods.
The proposed taxonomy and evidence framework
provide a basis for future comparative evaluations,
including controlled experiments across NLP tasks.

\section*{Use of AI assistance}
In preparing this manuscript, we used AI-assisted tools,
including Grammarly and ChatGPT, for proofreading,
grammar correction, and stylistic refinement.
OpenAI Codex was used to assist in developing the
HTML-based research register.
The authors take full responsibility for the
scientific content, analyses, and conclusions
presented in this work.

\bibliography{references}
\appendix

\section{Survey Methodology and Reproducibility}
\label{app:method}

We conducted a structured critical survey of SAE methods and their applications to language-model analysis
and NLP system construction, including relevant speech and
multimodal extensions. The literature search focused on
\emph{sparse autoencoders}, \emph{dictionary learning},
\emph{mechanistic interpretability}, \emph{language models},
\emph{feature circuits}, and \emph{activation steering}.
We consulted Google Scholar, Semantic Scholar, and the ACL
Anthology, covering publications primarily from 2024 to 2026
in ACL, EMNLP, ICLR, NeurIPS, ICML, and related venues,
including Findings papers, with a literature cutoff of 12 October 2026. Figure~\ref{fig:years} summarizes the included studies by reference-version year. Conference proceedings, PMLR,
arXiv, and OpenReview were also consulted to identify and
verify relevant studies and publication versions.

Of the 199 records collected, five duplicate or overlapping
version pairs were merged, yielding 194 unique candidates.
Following scope and source verification, \textbf{138 studies}
were included. Six additional methodological anchors and
four background works informed the discussion but were
excluded from the study-frequency counts. Each included
study was classified according to dictionary design,
feature operation, and NLP function, with relevant
experimental settings and source references retained.

The \reviewmaterials{} linked in the abstract provides the study records, coding guide, selection decisions, source-level evidence ledger, numerical extractions, and reproduction script as pages and individual files. Reported findings are distinguished from our interpretations.

\begin{figure}[tp]

\centering\begin{tikzpicture}[x=0.93cm,y=0.48cm,every node/.style={font=\small}]
\node at (0,.95) {2023};
\node at (1,.95) {2024};
\node at (2,.95) {2025};
\node at (3,.95) {2026};
\node[anchor=east] at (-.65,0) {Linguistic structure};
\fill[rep!8] (-0.46,-0.43) rectangle (0.46,0.43);
\node at (0,0) {0};
\fill[rep!8] (0.54,-0.43) rectangle (1.46,0.43);
\node at (1,0) {0};
\fill[rep!15] (1.54,-0.43) rectangle (2.46,0.43);
\node at (2,0) {5};
\fill[rep!12] (2.54,-0.43) rectangle (3.46,0.43);
\node at (3,0) {3};
\node[anchor=east] at (-.65,-1) {Dictionary design};
\fill[rep!9] (-0.46,-1.43) rectangle (0.46,-0.5700000000000001);
\node at (0,-1) {1};
\fill[rep!15] (0.54,-1.43) rectangle (1.46,-0.5700000000000001);
\node at (1,-1) {5};
\fill[rep!25] (1.54,-1.43) rectangle (2.46,-0.5700000000000001);
\node at (2,-1) {12};
\fill[rep!22] (2.54,-1.43) rectangle (3.46,-0.5700000000000001);
\node at (3,-1) {10};
\node[anchor=east] at (-.65,-2) {Text analysis};
\fill[rep!8] (-0.46,-2.43) rectangle (0.46,-1.57);
\node at (0,-2) {0};
\fill[rep!9] (0.54,-2.43) rectangle (1.46,-1.57);
\node at (1,-2) {1};
\fill[rep!15] (1.54,-2.43) rectangle (2.46,-1.57);
\node at (2,-2) {5};
\fill[rep!11] (2.54,-2.43) rectangle (3.46,-1.57);
\node at (3,-2) {2};
\node[anchor=east] at (-.65,-3) {Knowledge/retrieval};
\fill[rep!8] (-0.46,-3.43) rectangle (0.46,-2.57);
\node at (0,-3) {0};
\fill[rep!9] (0.54,-3.43) rectangle (1.46,-2.57);
\node at (1,-3) {1};
\fill[rep!15] (1.54,-3.43) rectangle (2.46,-2.57);
\node at (2,-3) {5};
\fill[rep!17] (2.54,-3.43) rectangle (3.46,-2.57);
\node at (3,-3) {6};
\node[anchor=east] at (-.65,-4) {Feature methodology};
\fill[rep!8] (-0.46,-4.43) rectangle (0.46,-3.57);
\node at (0,-4) {0};
\fill[rep!14] (0.54,-4.43) rectangle (1.46,-3.57);
\node at (1,-4) {4};
\fill[rep!27] (1.54,-4.43) rectangle (2.46,-3.57);
\node at (2,-4) {13};
\fill[rep!31] (2.54,-4.43) rectangle (3.46,-3.57);
\node at (3,-4) {16};
\node[anchor=east] at (-.65,-5) {Circuits/reasoning};
\fill[rep!8] (-0.46,-5.43) rectangle (0.46,-4.57);
\node at (0,-5) {0};
\fill[rep!15] (0.54,-5.43) rectangle (1.46,-4.57);
\node at (1,-5) {5};
\fill[rep!17] (1.54,-5.43) rectangle (2.46,-4.57);
\node at (2,-5) {6};
\fill[rep!21] (2.54,-5.43) rectangle (3.46,-4.57);
\node at (3,-5) {9};
\node[anchor=east] at (-.65,-6) {Safety/alignment};
\fill[rep!8] (-0.46,-6.43) rectangle (0.46,-5.57);
\node at (0,-6) {0};
\fill[rep!11] (0.54,-6.43) rectangle (1.46,-5.57);
\node at (1,-6) {2};
\fill[rep!17] (1.54,-6.43) rectangle (2.46,-5.57);
\node at (2,-6) {6};
\fill[rep!21] (2.54,-6.43) rectangle (3.46,-5.57);
\node at (3,-6) {9};
\node[anchor=east] at (-.65,-7) {Speech/multimodal};
\fill[rep!8] (-0.46,-7.43) rectangle (0.46,-6.57);
\node at (0,-7) {0};
\fill[rep!8] (0.54,-7.43) rectangle (1.46,-6.57);
\node at (1,-7) {0};
\fill[rep!9] (1.54,-7.43) rectangle (2.46,-6.57);
\node at (2,-7) {1};
\fill[rep!24] (2.54,-7.43) rectangle (3.46,-6.57);
\node at (3,-7) {11};
\end{tikzpicture}
\caption{Year of the selected reference version, after publication-metadata corrections. A 2024 preprint published in 2025 is counted in 2025. This is a cohort profile, not a publication-growth curve or first-appearance timeline.}
\label{fig:years}
\end{figure}

\section{Taxonomy Definitions and Cross-Dimensional Analysis}
\label{app:taxonomy}

\subsection{Operational definitions and multi-label rules}
For study $s$, $T(s)=(D_s,O_s,F_s;Z_s)$ records dictionary design, feature operation, and NLP function, with model/site, supervision, selection, intervention, and source-version qualifiers in $Z_s$. Table~\ref{tab:codebook} defines the labels and assignment boundaries. Labels describe source-supported contributions retained in the synthesis; multiple labels preserve overlap, while an absent secondary label denotes an unextracted contribution. The facets can interact and are not assumed statistically independent.

\begin{table*}[tp]
\centering
\footnotesize
\setlength{\tabcolsep}{4pt}
\renewcommand{\arraystretch}{1.12}

\begin{tabularx}{\textwidth}{
@{}
>{\raggedright\arraybackslash}p{0.06\textwidth}
>{\raggedright\arraybackslash}p{0.20\textwidth}
>{\raggedright\arraybackslash}X
>{\raggedright\arraybackslash}X
@{}
}
\toprule
\textbf{Code} &
\textbf{Category} &
\textbf{Assignment criterion} &
\textbf{Boundary or clarification} \\
\midrule

\multicolumn{4}{@{}l}{
\textbf{Dictionary design (D)}
} \\
\midrule

A & Adopted/comparative &
Uses or compares an existing dictionary; dictionary
construction is not the main contribution. &
Does not imply a new SAE architecture. \\

B & Budget/amplitude &
Changes sparsity allocation or separates feature
selection from activation magnitude. &
Using a TopK checkpoint alone does not qualify. \\

G & Geometry/structure &
Changes feature geometry, grouping, hierarchy,
signs, or relational constraints. &
A semantic label does not establish a
multidimensional representation. \\

C & Coverage/objective &
Changes training-data coverage or the objective
determining what is preserved. &
Task-specific supervision is recorded separately. \\

I & Inference/site/sharing &
Changes sparse-code inference, activation sites,
or parameter sharing. &
Choosing a layer for an application is not
necessarily a design contribution. \\

F & Functional/alternative &
Uses a functional surrogate, alternative basis,
or direct representation as a comparator. &
Transcoders model input--output mappings,
not same-state reconstruction. \\

\midrule
\multicolumn{4}{@{}l}{
\textbf{Feature operation (O)}
} \\
\midrule

D & Describe/characterize &
Analyzes features, representations,
reconstruction, or interpretability. &
A feature description does not establish
a natural computational mechanism. \\

P & Predict/diagnose &
Uses features to predict outcomes or
diagnose observed changes. &
Predictive association does not establish causation. \\

T & Trace dependencies &
Identifies or tests directed computational
dependencies or algorithmic relationships. &
Undirected coactivation alone is insufficient. \\

E & Edit/construct &
Modifies activations, inputs, parameters,
data, or SAE-based task systems. &
Ordinary SAE training is not itself
a system-construction contribution. \\

\midrule
\multicolumn{4}{@{}l}{
\textbf{NLP function (F)}
} \\
\midrule

L & Linguistic distinctions &
Investigates language identity, script,
grammar, lexical properties, or
language-linked speech factors. &
Using a language model alone does not
establish a linguistic contribution. \\

K & Knowledge/retrieval &
Studies retrieval, source reliance,
context use, or unsupported answers. &
Knowledge unlearning is primarily
behavioral regulation. \\

R & Reasoning/adaptation &
Studies reasoning, instruction following,
transfer, or model adaptation. &
Distinguish changes to weights,
inputs, and runtime activations. \\

S & Behavioral regulation &
Studies refusal, bias, unlearning,
adversarial behavior, or safety control. &
A narrow safety result does not
establish general safety. \\

A & Text analytics &
Studies classification, topic modeling,
text provenance, or text-based analysis. &
The target may be a downstream system,
not the original language model. \\

X & Speech/multimodal &
Studies language-linked acoustic,
visual, or cross-modal representations. &
Modality-specific findings may not
transfer to text-only models. \\

G & General/enabling &
Develops general SAE methods or tools
without a specific NLP application. &
Not a substitute for missing
methodological information. \\

\bottomrule
\end{tabularx}

\caption{Operational codebook for the proposed
three-facet taxonomy. Studies may receive multiple
codes within each dimension. Identical letters
can have different meanings across dimensions.
Assignments describe the contributions analyzed
in this survey, not study quality.}
\label{tab:codebook}
\end{table*}

The register serializes each facet as a set. Its compact notation is \emph{design / operation / function}; for example, A / DE / LK describes an adopted dictionary used to characterize and edit a language/retrieval representation. Codes within a set are alphabetized in printed tables; their order has no substantive meaning. A study can have several codes in a facet. An ambiguous label is retained only with a source-supported rationale; unresolved method details remain unassessed in the claim ledger. Qualifiers use explicit fields for model, site, dictionary, selection information, population, endpoint, comparator, version, and source locator. ``Unassessed'' means not extracted here, ``explicitly not evaluated'' requires a source statement, and ``not applicable'' requires a reason tied to the claim. None is a negative result.

Primary topics identify a study's central contribution; facet labels capture its additional roles. A multilingual retriever can therefore receive both knowledge/retrieval and linguistic-function labels. Table~\ref{tab:ambiguities} illustrates decisions where the representation, operation, or interpretation target would otherwise be conflated.

\begin{table*}[tp]
\centering
\footnotesize
\setlength{\tabcolsep}{4pt}
\renewcommand{\arraystretch}{1.15}

\begin{tabularx}{\textwidth}{
@{}
>{\raggedright\arraybackslash}p{0.19\textwidth}
>{\raggedright\arraybackslash}p{0.25\textwidth}
>{\raggedright\arraybackslash}X
>{\raggedright\arraybackslash}p{0.17\textwidth}
@{}
}
\toprule
\textbf{Ambiguous case} &
\textbf{Coding decision} &
\textbf{Why the distinction matters} &
\textbf{Examples} \\
\midrule

Hybrid SAE--classifier
&
Design: objective; operations:
describe, predict, construct;
function: text analytics.
&
An interpretable task-specific classifier
does not necessarily explain the
original language model.
&
DILA and task-specific classification
\citep{O056,O066}.
\\

\addlinespace[3pt]
Transcoder
&
Design: functional alternative;
operations: trace, characterize.
&
A transcoder approximates an
input--output mapping rather than
reconstructing the same activation.
&
Transcoder methods
\citep{O050,O100}.
\\

\addlinespace[3pt]
Sparse model diffing
&
Operations: describe, diagnose;
functions: adaptation, regulation.
&
Analyzing changes after an update
differs from predicting their effects
before training.
&
MNEME and transfer prediction
\citep{O022,N41}.
\\

\addlinespace[3pt]
SAE-informed data selection
&
Operation: construct;
function: adaptation or
multimodal extension.
&
Selecting training data changes
the resulting system; feature
analysis is a separate step.
&
IGDS and data filtering
\citep{N61,N31}.
\\

\addlinespace[3pt]
Feature-guided steering
&
Operation: edit; adopted
dictionary unless its design
is modified.
&
Performance gains may depend
on feature selection, optimization,
or gating rather than the
dictionary alone.
&
FGAA and REINS
\citep{O093,N71}.
\\

\addlinespace[3pt]
Coactivation modules
&
Operations: characterize, edit;
no automatic tracing label.
&
Coactivation and group interventions
do not establish directed
computational dependencies.
&
Coactivation modules
\citep{N47}.
\\

\addlinespace[3pt]
Direct comparators and
position papers
&
Record study role and genre
separately from taxonomy labels.
&
Empirical alternatives and
conceptual arguments provide
different forms of evidence.
&
Neuron circuits and
concept discovery
\citep{O105,O101}.
\\

\bottomrule
\end{tabularx}

\caption{Examples of ambiguous taxonomy assignments.
The coding decisions reflect the contributions examined
in this survey rather than every experiment reported
in each study.}
\label{tab:ambiguities}
\end{table*}

\subsection{Quantitative use of the taxonomy}
Figure~\ref{fig:cross} counts co-occurring labels in all 138 included studies. Each study contributes at most one count to a given cell; a multi-function or multi-operation study can contribute to several cells. Consequently, cell totals cannot be added to recover 138. The full register includes roles and genres so that a comparator, an extension, and a position paper are visible. The general-function category prevents forcing enabling research into an NLP application label.

\begin{figure*}[t]

\centering\begin{minipage}[t]{.49\textwidth}
\centering\textbf{Function $\times$ operation}\par\smallskip
\begin{tikzpicture}[x=0.7cm,y=0.54cm,every node/.style={font=\small}]
\node at (0,.95) {D};
\node at (1,.95) {P};
\node at (2,.95) {T};
\node at (3,.95) {E};
\node[anchor=east] at (-.65,0) {Linguistic};
\fill[rep!14] (-0.46,-0.43) rectangle (0.46,0.43);
\node at (0,0) {14};
\fill[rep!8] (0.54,-0.43) rectangle (1.46,0.43);
\node at (1,0) {0};
\fill[rep!8] (1.54,-0.43) rectangle (2.46,0.43);
\node at (2,0) {1};
\fill[rep!13] (2.54,-0.43) rectangle (3.46,0.43);
\node at (3,0) {13};
\node[anchor=east] at (-.65,-1) {Retrieval};
\fill[rep!12] (-0.46,-1.43) rectangle (0.46,-0.5700000000000001);
\node at (0,-1) {11};
\fill[rep!10] (0.54,-1.43) rectangle (1.46,-0.5700000000000001);
\node at (1,-1) {6};
\fill[rep!8] (1.54,-1.43) rectangle (2.46,-0.5700000000000001);
\node at (2,-1) {0};
\fill[rep!13] (2.54,-1.43) rectangle (3.46,-0.5700000000000001);
\node at (3,-1) {12};
\node[anchor=east] at (-.65,-2) {Reasoning/adaptation};
\fill[rep!14] (-0.46,-2.43) rectangle (0.46,-1.57);
\node at (0,-2) {15};
\fill[rep!9] (0.54,-2.43) rectangle (1.46,-1.57);
\node at (1,-2) {3};
\fill[rep!11] (1.54,-2.43) rectangle (2.46,-1.57);
\node at (2,-2) {8};
\fill[rep!14] (2.54,-2.43) rectangle (3.46,-1.57);
\node at (3,-2) {15};
\node[anchor=east] at (-.65,-3) {Regulation};
\fill[rep!14] (-0.46,-3.43) rectangle (0.46,-2.57);
\node at (0,-3) {14};
\fill[rep!10] (0.54,-3.43) rectangle (1.46,-2.57);
\node at (1,-3) {6};
\fill[rep!9] (1.54,-3.43) rectangle (2.46,-2.57);
\node at (2,-3) {2};
\fill[rep!14] (2.54,-3.43) rectangle (3.46,-2.57);
\node at (3,-3) {16};
\node[anchor=east] at (-.65,-4) {Text analytics};
\fill[rep!10] (-0.46,-4.43) rectangle (0.46,-3.57);
\node at (0,-4) {5};
\fill[rep!11] (0.54,-4.43) rectangle (1.46,-3.57);
\node at (1,-4) {7};
\fill[rep!8] (1.54,-4.43) rectangle (2.46,-3.57);
\node at (2,-4) {0};
\fill[rep!10] (2.54,-4.43) rectangle (3.46,-3.57);
\node at (3,-4) {5};
\node[anchor=east] at (-.65,-5) {Speech/multimodal};
\fill[rep!11] (-0.46,-5.43) rectangle (0.46,-4.57);
\node at (0,-5) {7};
\fill[rep!8] (0.54,-5.43) rectangle (1.46,-4.57);
\node at (1,-5) {1};
\fill[rep!8] (1.54,-5.43) rectangle (2.46,-4.57);
\node at (2,-5) {1};
\fill[rep!11] (2.54,-5.43) rectangle (3.46,-4.57);
\node at (3,-5) {7};
\node[anchor=east] at (-.65,-6) {General/enabling};
\fill[rep!31] (-0.46,-6.43) rectangle (0.46,-5.57);
\node at (0,-6) {58};
\fill[rep!8] (0.54,-6.43) rectangle (1.46,-5.57);
\node at (1,-6) {1};
\fill[rep!9] (1.54,-6.43) rectangle (2.46,-5.57);
\node at (2,-6) {2};
\fill[rep!14] (2.54,-6.43) rectangle (3.46,-5.57);
\node at (3,-6) {14};
\end{tikzpicture}
\end{minipage}\hfill
\begin{minipage}[t]{.49\textwidth}
\centering\textbf{Design $\times$ operation}\par\smallskip
\begin{tikzpicture}[x=0.7cm,y=0.54cm,every node/.style={font=\small}]
\node at (0,.95) {D};
\node at (1,.95) {P};
\node at (2,.95) {T};
\node at (3,.95) {E};
\node[anchor=east] at (-.65,0) {Adopted/comparative};
\fill[rep!31] (-0.46,-0.43) rectangle (0.46,0.43);
\node at (0,0) {64};
\fill[rep!14] (0.54,-0.43) rectangle (1.46,0.43);
\node at (1,0) {18};
\fill[rep!11] (1.54,-0.43) rectangle (2.46,0.43);
\node at (2,0) {7};
\fill[rep!28] (2.54,-0.43) rectangle (3.46,0.43);
\node at (3,0) {55};
\node[anchor=east] at (-.65,-1) {Budget/amplitude};
\fill[rep!11] (-0.46,-1.43) rectangle (0.46,-0.5700000000000001);
\node at (0,-1) {8};
\fill[rep!8] (0.54,-1.43) rectangle (1.46,-0.5700000000000001);
\node at (1,-1) {0};
\fill[rep!8] (1.54,-1.43) rectangle (2.46,-0.5700000000000001);
\node at (2,-1) {0};
\fill[rep!8] (2.54,-1.43) rectangle (3.46,-0.5700000000000001);
\node at (3,-1) {0};
\node[anchor=east] at (-.65,-2) {Geometry/structure};
\fill[rep!16] (-0.46,-2.43) rectangle (0.46,-1.57);
\node at (0,-2) {22};
\fill[rep!8] (0.54,-2.43) rectangle (1.46,-1.57);
\node at (1,-2) {1};
\fill[rep!8] (1.54,-2.43) rectangle (2.46,-1.57);
\node at (2,-2) {0};
\fill[rep!9] (2.54,-2.43) rectangle (3.46,-1.57);
\node at (3,-2) {3};
\node[anchor=east] at (-.65,-3) {Coverage/objective};
\fill[rep!12] (-0.46,-3.43) rectangle (0.46,-2.57);
\node at (0,-3) {10};
\fill[rep!9] (0.54,-3.43) rectangle (1.46,-2.57);
\node at (1,-3) {3};
\fill[rep!8] (1.54,-3.43) rectangle (2.46,-2.57);
\node at (2,-3) {0};
\fill[rep!11] (2.54,-3.43) rectangle (3.46,-2.57);
\node at (3,-3) {8};
\node[anchor=east] at (-.65,-4) {Inference/site/sharing};
\fill[rep!11] (-0.46,-4.43) rectangle (0.46,-3.57);
\node at (0,-4) {8};
\fill[rep!8] (0.54,-4.43) rectangle (1.46,-3.57);
\node at (1,-4) {0};
\fill[rep!8] (1.54,-4.43) rectangle (2.46,-3.57);
\node at (2,-4) {1};
\fill[rep!8] (2.54,-4.43) rectangle (3.46,-3.57);
\node at (3,-4) {1};
\node[anchor=east] at (-.65,-5) {Functional/alternative};
\fill[rep!11] (-0.46,-5.43) rectangle (0.46,-4.57);
\node at (0,-5) {9};
\fill[rep!8] (0.54,-5.43) rectangle (1.46,-4.57);
\node at (1,-5) {0};
\fill[rep!9] (1.54,-5.43) rectangle (2.46,-4.57);
\node at (2,-5) {4};
\fill[rep!11] (2.54,-5.43) rectangle (3.46,-4.57);
\node at (3,-5) {7};
\end{tikzpicture}
\end{minipage}
\caption{ Multi-label taxonomy of the 138 included studies. Left: operations by function. Right: design facets by operation. Cells count extracted contributions, not performed interventions or strong evidence. D/P/T/E denote characterize, predict, trace, and edit/construct. Full definitions and role/genre fields accompany the register.}
\label{fig:cross}
\end{figure*}

The linguistic-function count rises from eight primary-topic studies to 15 when secondary functions are retained. This is an observed property of our coding, not evidence that 15 independent causal mechanisms have been found. The map also separates prediction-focused analytics from dependency-tracing contributions and records the many interventions and system-construction studies that use an adopted dictionary. These patterns help choose case comparisons; they do not establish field-wide research prevalence, quality, or missing experiments.

\section{Technical Comparisons and Assumptions}
\label{app:technical}
Tables~\ref{tab:allocation} and~\ref{tab:geometry} expand RQ1 by separating allocation rules from geometric, structural, and inference assumptions, and by retaining each comparison's endpoint.
\subsection{Sparsity, scale, and allocation}
The $\ell_0$ quantity counts nonzero code coordinates.
For $m$ decoder columns $D_j$, reconstruction admits a scale ambiguity: $D_jz_j=(cD_j)(z_j/c)$. Without constraining or accounting for decoder norms, a model can reduce an activation penalty without changing reconstruction. Unit-norm decoders are one response; architectures can instead use scale-aware formulations. Comparisons should record input centering/normalization, decoder norms, nonnegativity or signed latents, and how intervention strength is translated into activation space. A unit coefficient in two dictionaries need not be an equal intervention.

With an $\ell_1$ objective, shrinking nonzero coefficients can lower loss even when the correct support is known. Gating or thresholding can mitigate this amplitude bias. Recovering the true concept dictionary is a different question, because correlated inputs, overcomplete alternatives, and inference error can preserve ambiguity after shrinkage is reduced. Binary codes remove one amplitude channel but change the rate--distortion trade-off \citep{O067,O103}. Reconstruction, sparsity, semantic discrimination, and downstream loss should therefore remain separate endpoints.

\begin{table*}[tp]
\centering
\footnotesize
\setlength{\tabcolsep}{4pt}
\renewcommand{\arraystretch}{1.15}

\begin{tabularx}{\textwidth}{
@{}
>{\raggedright\arraybackslash}p{0.14\textwidth}
>{\raggedright\arraybackslash}p{0.27\textwidth}
>{\raggedright\arraybackslash}p{0.29\textwidth}
>{\raggedright\arraybackslash}X
@{}
}
\toprule
\textbf{Method} &
\textbf{Allocation strategy} &
\textbf{Evidence and setting} &
\textbf{Implication for recoverability} \\
\midrule

TopK \citep{O081}
&
Selects a fixed number $k$ of
activations per input.
&
Evaluated through dictionary scaling,
reconstruction, and behavioral preservation.
&
A fixed budget may be insufficient for
complex inputs or excessive for simple ones.
\\

\addlinespace[3pt]
BatchTopK \citep{A05}
&
Selects $Bk$ activations across a batch;
uses an estimated threshold at inference.
&
GPT-2 Small and Gemma-2-2B.
Improved reconstruction over TopK;
mixed downstream cross-entropy
results against JumpReLU.
&
Reconstruction gains do not guarantee
better behavioral preservation or
semantic interpretability.
\\

\addlinespace[3pt]
Top-AFA \citep{A02}
&
Adapts the number of active features
using a quasi-orthogonality-based
magnitude approximation.
&
Theoretical approximation bounds
and GPT-2 experiments on OpenWebText.
&
Adaptive sparsity can follow
geometric assumptions without a
learned complexity predictor.
\\

\addlinespace[3pt]
AdaptiveK \citep{N58}
&
Predicts input complexity to determine
the sparsity budget.
&
Empirical evaluations of
complexity-dependent allocation
and downstream behavior.
&
Adaptive allocation does not by
itself establish recovery of
meaningful semantic features.
\\

\addlinespace[3pt]
GBA \citep{N40}
&
Adapts feature activation frequencies
across groups under specified
coverage assumptions.
&
Synthetic feature-recovery experiments;
performance depends on frequency
coverage (Appendix~\ref{app:quant}).
&
Recovery depends on coverage
assumptions; synthetic results
do not establish NLP generalization.
\\

\bottomrule
\end{tabularx}

\caption{Comparison of sparsity-allocation methods
and their implications for feature recoverability.
The methods differ in their allocation rules,
assumptions, and evaluation settings.
Improvements in reconstruction or synthetic recovery
do not necessarily imply better semantic
interpretability or downstream performance.}
\label{tab:allocation}
\end{table*}

\begin{table*}[tp]
\centering
\footnotesize
\setlength{\tabcolsep}{4pt}
\renewcommand{\arraystretch}{1.15}

\begin{tabularx}{\textwidth}{
@{}
>{\raggedright\arraybackslash}p{0.20\textwidth}
>{\raggedright\arraybackslash}p{0.27\textwidth}
>{\raggedright\arraybackslash}p{0.24\textwidth}
>{\raggedright\arraybackslash}X
@{}
}
\toprule
\textbf{Method / design} &
\textbf{Key assumption or change} &
\textbf{Reported evidence} &
\textbf{Implication or limitation} \\
\midrule

Gated / JumpReLU SAEs
\citep{O087,O053}
&
Separate feature selection from
activation magnitude, or use
learned activation thresholds.
&
Improvements in sparsity and
reconstruction under the
respective training objectives.
&
Reducing shrinkage does not
guarantee recovery of the
intended semantic features.
\\

\addlinespace[3pt]
Signed / subspace features
\citep{N38,N20,O049}
&
Represent concepts using signed
coordinates or multidimensional
subspaces instead of single
nonnegative features.
&
Investigations of feature polarity,
distributed concepts, and
multidimensional geometry.
&
Comparisons must account for
feature dimensionality and
representation capacity.
\\

\addlinespace[3pt]
Hierarchy / composition
\citep{O114,N13,N73}
&
Introduce nested dictionaries,
parent--child dependencies,
or compositional feature structures.
&
Methods for representing
multiple feature scales,
hierarchies, or compositions.
&
Learned hierarchies do not
necessarily correspond to
linguistic structures or
causal circuits.
\\

\addlinespace[3pt]
Weighted SAE (WSAE)
\citep{N30}
&
Analyzes recovery under
specified geometric, sparsity,
and tied-weight assumptions.
&
Theoretical recovery results
under an idealized
generative model.
&
Recovery guarantees do not
automatically extend to
real language-model activations.
\\

\addlinespace[3pt]
Geometry-oriented theory
\citep{N01}
&
Relates feature recovery to
concept geometry and the
reconstruction model.
&
Identifies conditions under
which particular concept
structures are recoverable.
&
Recovering a geometric
structure does not establish
its linguistic meaning.
\\

\addlinespace[3pt]
Local Amortized SAE
(LocA-SAE) \citep{N35}
&
Addresses the gap between
shared-encoder predictions
and per-input sparse inference.
&
Gemma-2-2B experiments
with iterative inference,
including a 200-step ISTA reference.
&
Improved inference does not
establish recovery of
ground-truth semantic features.
\\

\bottomrule
\end{tabularx}

\caption{Comparison of SAE design assumptions and their
implications for feature recoverability. The methods
address different aspects of representation learning,
including sparsity, geometry, structure, and inference.
Their reported results should be interpreted under
the corresponding assumptions and evaluation settings.}
\label{tab:geometry}
\end{table*}

\subsection{Identifiability and the object represented}
The \emph{linear representation hypothesis} concerns representing a concept through a direction or low-dimensional linear structure \citep{O044}. \emph{Superposition} concerns encoding more features than available dimensions using overlapping directions. Neither assumption by itself guarantees that a particular learned dictionary uniquely recovers the intended concepts.

A low reconstruction error means the selected codes reproduce activations under a metric. It does not guarantee that a unique dictionary is identifiable from that distribution. Permutation/scaling symmetries are elementary ambiguities; correlated or hierarchical factors can support substantively different decompositions. Ground-truth recovery is measurable in synthetic settings with a known generating dictionary. In a natural LM, semantic tests, controlled tasks, and functional comparisons provide partial constraints, rather than direct access to an agreed ground-truth feature inventory.

\emph{Splitting} represents a nominal concept through multiple latents, sometimes at different granularities. \emph{Absorption} occurs when a specialized feature also carries a more general property, so absence of the nominal general feature's activation does not imply absence of that information \citep{N03}. \emph{Composition} concerns how multiple factors combine to represent or compute a target; KronSAE's factorized interactions implement a different assumption from either phenomenon \citep{N73}. Ablating one split coordinate may underestimate distributed involvement; ablating an absorbing coordinate can affect several semantic properties. These are reasons to inspect feature sets and their side effects, not to assume every feature group is a causal mechanism.

A multidimensional concept occupies a subspace or requires several coordinates \citep{O049,N20}. A concept manifold introduces continuous geometry that may be represented by global spanning features or local tiling \citep{O109}. Matryoshka dictionaries couple nested reconstruction scales \citep{O114}; hierarchical gating models explicit parent--child dependence \citep{N13}; cascaded SAE++ learns higher-level concepts over decoder directions \citep{N08}. These differences determine what an intervention on a node, group, prefix, or subspace means. The rate--distortion--polysemanticity model further shows why co-occurrence can change the preferred decomposition \citep{O112}.

\subsection{Sequence, coverage, and computational site}
Temporal smoothness and predictable-context decomposition are different hypotheses about sequences \citep{O069,N27}. The former favors slowly changing higher-level factors; the latter separates predictable information from novelty. Neither supplies semantic ground truth: repeated lexical forms can be temporally stable, and a linguistically important contrast can be predictable. A discourse experiment should independently vary semantic continuity, lexical repetition, and information novelty, measuring both feature organization and intervention effects.

Coverage is an independent source of failure. A concept absent or extremely rare in the training activations cannot be recovered as reliably as one repeatedly represented in appropriate contexts. Specialized SAEs change sampling and risk weighting \citep{O031}; multilingual coverage adds language, script, tokenization, and construction-frequency imbalances. General-purpose and specialized dictionaries should be compared with their different data and supervision budgets visible. An apparent architectural advantage can otherwise be a coverage advantage.

Residual-stream SAEs characterize a mixed running state; attention-output SAEs expose context-dependent writes; MLP-output SAEs characterize another update to that state. A transcoder learns a sparse approximation to an input--output function, creating a different candidate causal object. A pooled-embedding SAE for retrieval collapses token position and belongs to a different representation level. Speech frames, visual tokens, and diffusion timestep/mask position introduce further indexing variables.

\section{Evidence Framework and Causal Counterexample}
\label{app:evidence}
Table~\ref{tab:rubric} relates each claim to its evidential requirements; Table~\ref{tab:worked_evidence} applies those distinctions to published findings. A semantic description requires held-out contrasts and discrimination from neighboring features. Predictive and control claims require matched supervision, selection, and baselines. A mechanistic claim additionally concerns the computation used by the unmodified model. Evidence is assessed for the stated proposition and setting, rather than assigned as a single paper-level score. Full extraction fields and assessment rules accompany the register (Appendix~\ref{app:method}).

\begin{table*}[tp]
\centering
\footnotesize
\setlength{\tabcolsep}{4pt}
\renewcommand{\arraystretch}{1.15}

\begin{tabularx}{\textwidth}{
@{}
>{\raggedright\arraybackslash}p{0.20\textwidth}
>{\raggedright\arraybackslash}p{0.27\textwidth}
>{\raggedright\arraybackslash}p{0.25\textwidth}
>{\raggedright\arraybackslash}X
@{}
}
\toprule
\textbf{Scientific claim} &
\textbf{Reported evidence} &
\textbf{What can be concluded} &
\textbf{Further validation} \\
\midrule

Features represent
abstract reasoning
&
Reasoning-related activations;
cue-injection and
meaning-preserving tests
\citep{O076,O072}.
&
Activation patterns alone do
not establish abstract reasoning;
counterexamples can weaken
the interpretation.
&
Test semantic contrasts
independently of
control utility.
\\

\addlinespace[3pt]
Feature descriptions
identify distinct features
&
Descriptions predict activations
but may also fit unrelated
features \citep{O111}.
&
A plausible description does
not necessarily distinguish
one feature from another.
&
Use hard negatives and
neighboring-feature
discrimination.
\\

\addlinespace[3pt]
Successful patching
reveals a mechanism
&
A patch can reproduce an
output through disconnected
and dormant components
\citep{A01}.
&
A causal intervention need
not recover the model's
natural computation.
&
Test natural pathways and
component-specific effects.
\\

\addlinespace[3pt]
Sparse circuits explain
grammatical behavior
&
Feature/error circuits with
controlled interventions on
agreement tasks
\citep{O083}.
&
Provides evidence for a
mechanistic role within
the tested setting.
&
Test transfer across tasks,
contexts, and distributions.
\\

\addlinespace[3pt]
SAEs improve
classification
&
Positive task-specific and
pooled representations;
mixed generic probing
results \citep{O021,O066,O063}.
&
Benefits depend on the
representation, supervision,
and classification pipeline.
&
Match feature selection,
pooling, model capacity,
and baselines.
\\

\addlinespace[3pt]
SAEs enable safe
knowledge removal
&
Forgetting results under
different populations,
controllers, and retention
tests \citep{O059,O092,N59}.
&
Unmatched results do not
establish a general
forgetting--retention
advantage.
&
Compare selected populations,
retention, intervention type,
and relearning.
\\

\bottomrule
\end{tabularx}

\caption{Examples of claim-dependent evidence in SAE
research. Each case distinguishes reported findings
from the conclusions they support and identifies
additional validation needed for broader claims.
Assessments concern specific claims and evaluation
settings, not the overall quality of the cited studies.}
\label{tab:worked_evidence}
\end{table*}

\subsection{Why causal success can mislead: an explicit example}
\label{app:illusion}
\begin{figure}[tp]

\centering\begin{tikzpicture}[x=1cm,y=1cm,>=Stealth,every node/.style={font=\small},box/.style={draw,rounded corners,align=center,text width=2.5cm,inner sep=4pt}]
\node[box] (read) at (0,0) {Coordinate 1\\correlated, ignored\\$h_{A,1}=2$};
\node[box] (natural) at (4,0) {Coordinate 3\\natural route\\$h_{A,3}=1$};
\node[box] (write) at (0,-1.85) {Coordinate 2\\dormant, sensitive\\$h_{B,2}=0\to1$};
\node[box] (out) at (4,-1.85) {Output\\$g(h)=h_2+h_3$\\patched value $=1$};
\draw[->,thick,dashed] (read)--node[left,align=right,font=\footnotesize]{patch} (write);
\draw[->,thick] (write)--(out);
\draw[->,thick] (natural)--node[right,font=\footnotesize]{natural} (out);
\end{tikzpicture}
\caption{A successful patch can create an alternative route. Solid labels describe the natural computation in the illustrative example; the patch transfers input information from the ignored coordinate into the dormant output-sensitive coordinate.}
\label{fig:illusion}
\end{figure}
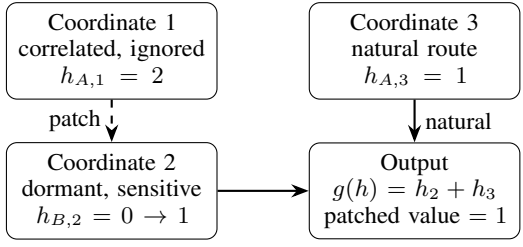

Figure~\ref{fig:illusion} illustrates how a successful intervention can bypass the natural computation. Following the disconnected/dormant decomposition of \citet{A01}, consider a pedagogical three-coordinate instantiation. The model naturally has $h_A=(2,0,1)$ and $h_B=(0,0,0)$, with downstream output $g(h)=h_2+h_3$. Coordinate 1 distinguishes inputs but is ignored downstream; coordinate 2 can change the output but is naturally dormant here; coordinate 3 carries the natural output difference.

Let $v=(1,1,0)/\sqrt 2$. Patching the projection from $A$ into $B$ gives
\begin{align}
 h'_B &= h_B+v\,v^\top(h_A-h_B)=(1,1,0),\\
 g(h_A)&=1,\qquad g(h_B)=0,\qquad g(h'_B)=1.
\end{align}
The patch passes an output-substitution test while leaving coordinate 3 unchanged. Its read signal came from a disconnected coordinate, and its write effect used the dormant coordinate. An encoder proportional to $[1,1,0]$ can therefore track a useful input contrast without identifying the route naturally responsible for the output. This illustration is our explicit instantiation of the published argument, not a new experiment or a claim that all trained SAEs exhibit the failure.

The feature claim should be weakened at the natural-pathway check, even if its activation label and output intervention both look correct. Testing the disconnected and dormant components separately exposes the mismatch. This answers the question of whether causal validation is sufficient: an intervention is causal as a manipulation, but its success need not validate the proposed interpretation of the unmodified computation.

\subsection{Reconstruction, distribution shift, and edges}
Equation~\ref{eq:edit} preserves the original reconstruction residual. A null edit $z'=z$ then returns the original activation, while replacing it with the decoded reconstruction can alter the model before any intended edit occurs. This distinguishes reconstruction damage from the effect of the feature delta. It does not ensure that the edited state was naturally reachable. Large coefficients, incompatible coactivation, or simultaneous edits across sites can create off-manifold states.

Matched-norm random directions, naturally occurring replacement values, coefficient sweeps, retention tasks, and token-level effects probe different confounds. They should be selected for the claim rather than applied as an undifferentiated checklist. A failed off-target test may weaken specificity while leaving a target causal effect intact. A null necessity result can reflect redundant computation; it does not by itself prove irrelevance. Sufficiency is likewise conditional on the receiving context and intervention convention.

An edge in a circuit diagram should state its construction: activation correlation, geometric matching, gradient-based attribution, or intervention-confirmed dependence. These are not interchangeable. Attribution can prioritize candidate edges efficiently, while controlled removal/restoration and counterfactual tests assess whether the selected path supports the task. Transcoder reconstruction of a component function and SAE reconstruction of one state have different error terms; a circuit comparison must retain those differences.

\section{Deep Comparative NLP Cases}
\label{app:cases}
These cases extend RQ3 by comparing representations, objectives, and intervention pipelines. Numerical contrasts are examined in Appendix~\ref{app:quant}; selected source observations and their interpretations are documented in Appendix~\ref{app:ledger}.
\subsection{Linguistic identity, grammatical variables, and transfer}
The multilingual literature moves through several distinct objects. Language-associated groups \citep{O007} concern broad selectivity and combined interventions. LinguaLens constructs counterfactual contrasts and tests ablation/enhancement on Llama-3.1-8B with OpenSAE checkpoints; its causal estimates depend on the selected examples and LLM-judged outcome \citep{O019}. TenseLoC distinguishes locating an attention-output bottleneck from steering selected residual-stream SAE features \citep{O032}. Script-controlled analysis demonstrates why broad language associations may track writing systems; its raw-neuron interventions belong to a direct-comparator role \citep{N51}.

The comparison suggests an ordered set of questions, not an evidence hierarchy across papers. First identify what varies: language identity, script, tense, meaning, or discourse role. Then hold matched semantic content where possible, use minimal grammatical contrasts, and test paraphrase or transliteration when the claim predicts invariance. Finally assess whether a matched feature preserves the predicted intervention effect across languages. Vector similarity alone does not establish that equivalence.

Evidence from Italian models \citep{O013} and the studied multilingual sets establishes broader coverage than English-only discovery. It does not justify an extrapolation to all low-resource languages or morphologically diverse systems. Tokenization, script, corpus frequency, and syntactic constructions can each change the learned decomposition. A useful next comparison samples these factors explicitly and reports per-language semantic and functional outcomes rather than only a macro-average.

\subsection{Language identity as a target and as a nuisance}\label{app:language_identity}
LangSAE uses a frozen multilingual-e5-large encoder, mean-pooled 1,024-dimensional representations, and a TopK-style sparse dictionary; it removes language-associated components before normalized retrieval scoring \citep{N64}. Its training construction draws on mMARCO and MIRACL in ten languages. The question is whether the edited representation preserves semantic query--document relations while reducing an unwanted language cue.

CLaS-Bench instead studies output-language control in an autoregressive instruction model \citep{N67}. A language cue that is undesirable for language-invariant retrieval is now the target signal. The source compares direct residual methods, probes, and SAE-derived directions under method-specific settings. Its mean score combines language fidelity and response relevance, and SAE comparisons are limited to layers with available checkpoints. Consequently, the observed direct-space advantage does not isolate dictionary quality from layer availability and selection procedure. The common term ``language feature'' conceals these opposite objectives and different representation sites.

\subsection{Prediction versus construction in sparse classification}\label{app:classification}
\citet{O063} evaluate sparse probes on Gemma-2-9B with JumpReLU GemmaScope dictionaries, with a Llama-3.1-8B/TopK replication. They examine last-token and multi-token max-pooled representations, attention-pooled dense baselines, and feature binarization. \citet{O021} use frozen Gemma-family models and public dictionaries, vary pooling and binarization, and train logistic regression with five-fold cross-validation. Their neural baseline uses last-token hidden states. \citet{O066} instead align representations to classification tasks; their encoder-only and decoder-only models also use different extraction sites.

\textbf{Source observation:} the within-study baseline comparison in \citet{O063} reduces the apparent pooled-SAE advantage when attention-pooled dense probes are available. This provides direct evidence that baseline construction matters in that protocol. \textbf{Survey interpretation:} positive classification results and limited generic-probing gains need not contradict each other because the representation, aggregation, selected features, classifier, and baseline differ. It would be inaccurate to attribute the difference simply to one paper using pooling and the other omitting it. Nor does this comparison isolate the cause of every difference between \citet{O021,O063,O066}. A matched study should fix model, layer, dataset split, label and tuning budgets while varying pooling, feature selection, task alignment, and head capacity.

\subsection{RAG: detection, source arbitration, and oracle repair}\label{app:rag}
Table~\ref{tab:rag} summarizes the stage and information requirements underlying these comparisons.
Dense embedding decomposition \citep{O054} and retrieval-oriented objectives \citep{O034,N25} can produce controllable representations. Query-expansion feature discovery selects 2,000 pairs that improve MRR@10 \citep{N80}. The source also analyzes helpful, neutral, and harmful expansions (their Appendix B.2); this broader diagnostic analysis does not remove the success-conditioned selection of steering latents. SAFE integrates feature signals into query enrichment and subsequent generation \citep{O029}. All of these are system-level contributions whose SAE-specific benefit requires a pipeline ablation.

HalluSAE and RAGLens predict hallucination-related outcomes \citep{O090,N32}. RAGLens evaluates Llama2-7B/13B on RAGTruth and Dolly and tests additional model/dataset combinations; its detector and feedback stage are separate evaluations. Re-encoding through another backbone is not evidence that identical latent indices or meanings transfer. Sparse-latent source arbitration in Llama-3.1-8B \citep{O006} and SPARE \citep{O033} instead changes reliance on context or internal knowledge. A proper correctness analysis must cross the intervention with whether each source is trustworthy.

LatDefuse selects distractor-vulnerable BioASQ, Natural Questions, and PopQA cases, then optimizes latent activations toward the known correct token at the first answer divergence \citep{N55}. Both the host model and SAE remain frozen. The selected population contains approximately 255 instances per model, with gold context between two hard distractors. Its DCD/MDR decoding baselines also use oracle information---gold versus distractor document roles---but this is different from the correct-token optimization target. \textbf{Source observation:} optimized edits restore answers in this selected setting. \textbf{Survey interpretation:} this establishes conditional repair capacity, not autonomous error detection or correction; post-hoc feature labels do not uniquely identify a natural semantic mechanism. The source acknowledges supervision and semantic-interpretation limits. Its Llama naming remains inconsistent: its Section 4.1 and footnote 2 name Llama-3-8B, while its Table 1 and SAE checkpoint name Llama-3.1-8B. We retain that unresolved discrepancy in the ledger and use the unambiguous Gemma-2-2B setting for concrete model examples.

\subsection{Reasoning and instruction: labels, cues, and supervision}
Reasoning steering \citep{O076,O104} and reasoning-feature falsification \citep{O072} concern different propositions. A feature can help induce a strategy while responding to a phrase rather than an abstract operation. Meaning-preserving rewrites and cue insertion test the semantic label; controlled reasoning tasks and targeted counterfactuals test the computational interpretation. Temporal-difference analysis supplies explicit reward/value/error variables in a specified in-context reinforcement-learning task \citep{O082}; it should not be described as a temporal SAE architecture or as general evidence about chain-of-thought reasoning.

For instruction following, generated compliant/violating narrative contrasts \citep{O017}, prompt-variation feature selection \citep{O020}, instruction-associated latents \citep{O062}, and labeled demonstration utility \citep{N17} provide different information to the method. A comparison that grants a known target to one selector and only an unsupervised label to another does not isolate representation quality. In sentiment circuits, the same intervention has markedly different gains across prompt templates \citep{N49}; this motivates reporting the prompt baseline as part of the method's information and evaluation context.

CoCoMix changes pretraining by predicting and interleaving SAE-derived concepts \citep{N23}; IGDS selects fine-tuning data using feature recall and intervention-based relevance \citep{N61}. Transfer prediction uses pre-update signals \citep{N41}, whereas MNEME diagnoses changes after an update \citep{O022}. These stages have distinct prediction targets and access to labels or updated weights. Input optimization and inference-time activation edits are not weight adaptation. Diffusion-language work additionally conditions its feature interpretation on timestep and mask state \citep{O106}.

\subsection{Safety and unlearning: location, conditionality, persistence}
Feature discovery, adversarial detection, and control should be extracted as separate claims. Constant refusal amplification has an observable target/benign-utility trade-off \citep{O060}. REINS uses harmful-feature suppression, refusal enhancement, and gating on Qwen3.5 base models \citep{N71}; graph-regularized steering also combines structured directions with a runtime decision \citep{N12}. These are composite controllers, so an improvement over static steering need not be an improvement in feature semantics.

Early SAE unlearning evaluates selected knowledge and has mixed multi-topic results \citep{O059}. Dynamic SAE Guardrails use Gemma-2-2B-it with a layer-3 residual SAE and input-dependent feature statistics \citep{O092}. SSPU and CRISP change parameters using SAE-derived constraints \citep{O018,N59}. The former runtime gate can avoid acting on unrelated inputs; persistent parameter changes eliminate that particular runtime dependency but create a different forgetting/retention and relearning problem. We compare where the change resides and what information selects it; a pooled ranking across unmatched benchmark denominators would hide these differences.

Reward modeling \citep{O077}, low-rank adaptation \citep{N60}, and coupled constraints during fine-tuning \citep{N66} construct interpretable or constrained training systems. Emergent-misalignment geometry \citep{N50}, medical-bias steering \citep{N53}, and fine-grained safety resources \citep{N57} add domain-specific diagnostic and control targets. Backdoor analyses find different roles for trigger detection, propagation, and output control across attack constructions and sites \citep{N77,N82}. These results motivate reporting the attack and benign population, not extrapolating a narrow diagnostic success to general safety.

\subsection{Text analytics, speech, and multimodal extensions}
Company similarity \citep{O005}, artificial-text detection \citep{O026}, medical coding \citep{O056}, classification \citep{O066}, and topic modeling \citep{N68} use sparse features to construct or explain task systems. Watermark sampling \citep{N05} and text-based confounding adjustment \citep{N74} have still different targets. In the latter, causal adjustment concerns an external data-generating question, not necessarily the language model's internal causal mechanism. These studies remain in the map to mark the boundary between interpretable system construction and source-model explanation.

Whisper analysis operates on acoustic encoder frames rather than text-token residual states \citep{O110}. Cross-modal feature heterogeneity, energy alignment, and group-sparse dictionaries test how text and visual representations share or separate concepts \citep{N18,N34,N36}. Hierarchical CLIP features and cascaded SAE++ introduce further structural objects \citep{O113,N08}; multimodal circuit tracing models input--output dependencies using transcoders \citep{O107}. A semantic label can be shared while directions or causal roles differ, so text-only geometric conclusions do not transfer automatically.

Other extensions construct systems: sparse-feature distribution matching selects image-caption training data \citep{N31}, concept coverage guides token pruning \citep{N63}, and caption-conditioned masks alter visual representations for retrieval \citep{N81}. Representation analysis motivates multimodal retrieval realignment \citep{N56}; text-derived visual steering also contains non-SAE alternatives, whose gains must not be attributed wholly to SAEs \citep{N65}. These cases test representation portability and objective dependence, while their specific accuracy or retrieval outcomes remain modality- and protocol-dependent.

\section{Secondary Analyses of Published Results}
\label{app:quant}
\begin{figure*}[tp]

\centering\begin{tikzpicture}
\begin{axis}[name=a,width=7.7cm,height=4.8cm,ymin=0,ymax=100,xmin=.7,xmax=4.3,xtick={1,2,3,4},xticklabels={T1,T2,T3,T4},ytick={0,25,50,75,100},ylabel={SST-2 F1 (\%)},xlabel={Prompt template},title={\textbf{(a) Prompt-conditioned gains}},font=\small,grid=major,grid style={black!10},legend style={font=\footnotesize,at={(.97,.03)},anchor=south east,draw=none,fill=white},legend cell align=left]
\addplot[color=black!65,mark=square*,thick] table[x=template,y=baseline,col sep=comma]{data/sentiment.csv};
\addlegendentry{Unmodified model}
\addplot[color=rep,mark=*,thick] table[x=template,y=intervention,col sep=comma]{data/sentiment.csv};
\addlegendentry{Circuit intervention}
\end{axis}
\begin{axis}[at={(a.east)},anchor=west,xshift=1.25cm,width=7.4cm,height=4.8cm,ybar,ymin=0,ymax=100,xmin=.5,xmax=4.5,xtick={1,2,3,4},xticklabels={English\\prompt,Target\\prompt,Residual\\DiffMean,SAE\\DiffMean},xticklabel style={font=\footnotesize,align=center},ytick={0,25,50,75,100},ylabel={Mean LSS (0--100)},title={\textbf{(b) A direct-space alternative}},font=\small,grid=major,grid style={black!10},bar width=18pt,nodes near coords,nodes near coords style={font=\footnotesize,anchor=south}]
\addplot[fill=ctrl!70,draw=ctrl] table[x=index,y=lss,col sep=comma]{data/language_steering_plot.csv};
\end{axis}
\end{tikzpicture}
\caption{Published within-study results, replotted for two distinct comparisons. Left: Gemma-2-2B SST-2 F1 under four templates and sentiment-circuit amplification \citep{N49}; three-run means are reported without dispersion in the extracted table. Right: mean CLaS-Bench language-steering score for Llama-3.1-8B-Instruct over 32 languages \citep{N67}. Different tasks, metrics, and selection protocols prohibit a cross-panel ranking.}
\label{fig:quant}
\end{figure*}
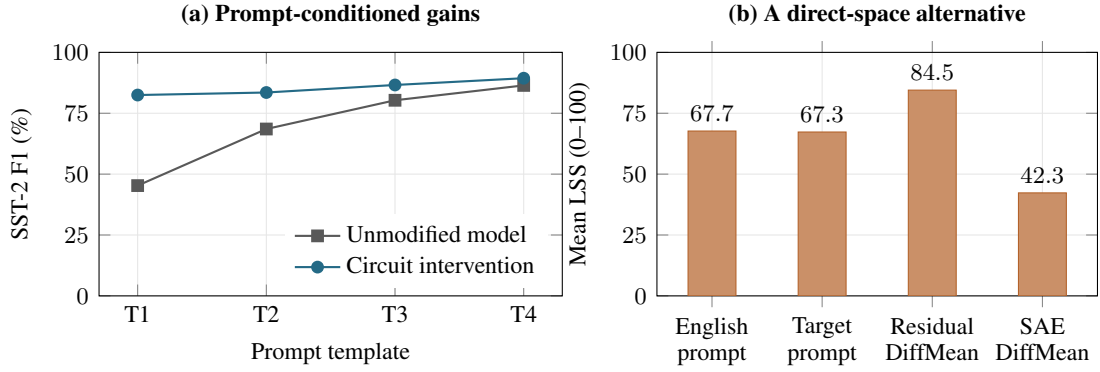

The numerical panels were selected because each source supplies a within-study contrast relevant to a different part of the argument: changing the prompt while holding the intervention family fixed, and changing the representation/control method under a language-steering benchmark. They are illustrative mechanism-of-comparison cases, not a representative meta-analysis. The source results are replotted in Figure~\ref{fig:quant}; our additions are subtraction of percentage-point differences and their interpretation in the taxonomy. We run no new model experiments.

\subsection{Sentiment and prompt dependence}\label{app:sentiment}
The left panel of Figure~\ref{fig:quant} contrasts baseline and intervention scores under the four prompts.
\citet{N49} report baseline F1 of 45.29, 68.53, 80.33, and 86.43 for templates T1--T4, and intervention F1 of 82.48, 83.52, 86.61, and 89.39. Their differences are 37.19, 14.99, 6.28, and 2.96 \emph{percentage points}. These are absolute differences, not relative percentage improvements.

The reported results concern the Gemma-2-2B setting; Llama-3.1-8B results are separate. GemmaScope 16k dictionaries with average $\ell_0$ nearest 100 cover residual, MLP, and attention sites. The sentiment circuit is derived from SST-2, with node/edge thresholds 0.073/0.007. Strength is calibrated on contrastive validation data to target a logit-margin gain of 2.0 and is capped at 3. Evaluation selects the label using first output-token logits, without free-generation decoding. The source's ``zero-shot'' label refers to prompting; it does not imply unsupervised circuit discovery or strength selection. \textbf{Observation:} the three-run mean F1 differences vary across T1--T4. \textbf{Interpretation:} useful gains are prompt-conditioned in this setting; differences in baseline quality and remaining headroom are not isolated causal explanations. The extracted results supply no dispersion, so no error bars or significance tests are added.

\subsection{Language steering and direct-space alternatives}\label{app:steering}
The right panel of Figure~\ref{fig:quant} complements the objective comparison in Appendix~\ref{app:language_identity} with a within-benchmark contrast between intervention pipelines.
\citet{N67} report mean language-steering scores of 67.7 for English prompts, 67.3 for target-language prompts, 84.5 for residual difference-in-means, 48.6 for a linear probe, 15.1 for PCA, 42.3 for SAE difference-in-means, 23.6 for LDA, and 80.1 for LAPE. Figure~\ref{fig:quant} (right) selects the two prompt baselines and two difference-in-means representations; the numerical-analysis page of the online register retains all eight values. The residual-minus-SAE difference is 42.2 score points.

The language-steering score is the harmonic mean $\mathrm{LSS}=2\,\mathrm{LFS}\,\mathrm{OR}/(\mathrm{LFS}+\mathrm{OR})$, reported on a 0--100 scale: LFS is the target-language success fraction measured with FastText language identification; OR is normalized 0/1/2 relevance scored by Qwen-3-8B. The reported scores average over 32 languages on Llama-3.1-8B-Instruct. The selected settings use residual DiffMean at layer 20 and SAE DiffMean at layer 25, both with strength 5, based on the source's ablations. We retain these reported settings; the conflicting prose setting is recorded in Case~\ref{case:E10}. SAE checkpoints exist only at layers 4, 12, 18, 20, and 25. The SAE intervention adds the reconstruction-error residual back after decoding; its gap cannot simply be attributed to replacing the hidden state with a lossy reconstruction. \textbf{Observation:} residual DiffMean has the higher reported score. \textbf{Interpretation:} this supports the complete selected pipeline in this benchmark, without isolating reconstruction error, language coverage, or dictionary quality as its cause. Neither equal layer coverage/compute nor an independent held-out tuning protocol is inferred. No uncertainty or significance estimate is reconstructed from the means.

\subsection{Selected denominators in unlearning and repair}
\citet{O059} evaluate WMDP-Bio questions answered correctly under all 24 answer-option permutations: 172 of 1,273 for Gemma-2B-it and 522 for Gemma-2-2B-it. This targets reliably available knowledge rather than the entire benchmark population. LatDefuse similarly conditions on distractor-vulnerable cases and a known correct token \citep{N55}. Their rates therefore cannot be pooled with full-benchmark or autonomous-system accuracy. We retain the selected denominators as part of the claim rather than treating them as generic limitations detached from the comparison.

\subsection{Capacity allocation and synthetic recovery}\label{app:recovery}
This comparison expands the coverage-sensitive result in Table~\ref{tab:allocation}; the online evidence ledger supplies the source locator for Case~\ref{case:E02}.
\citet{N40} use 128 generating features in 42 dimensions and an 8,192-unit SAE; half the normalized samples sum three features and half sum 20. The source clarifies that \emph{sample sparsity varies but marginal feature frequencies are uniform}, $f=(3+20)/(2\cdot128)\approx0.09$. Full-coverage GBA uses ten target-frequency groups spanning 0.001--0.5; misspecified coverage uses ten groups spanning 0.001--0.01, excluding $f$. At alignment threshold 0.9, TopK with $K=20,30,50$ recovers 98.4\%, 24.2\%, and 23.4\%; full/misspecified GBA recover 100\%/3.9\%. Results at the 0.8 threshold differ. \textbf{Observation:} recovery changes with these settings. \textbf{Interpretation:} this is sensitivity to sparsity allocation and target-frequency coverage in a known synthetic model, not evidence about long-tail linguistic recovery or NLP accuracy.

\section{Selected Source-Level Evidence Cases}
\label{app:ledger}
The six cases below document the observations underlying the recovery, semantic-validation, and NLP comparisons. Original case identifiers link them to the complete 28-case ledger in the online register (Appendix~\ref{app:method}), which retains detailed extraction fields, source locators, and source-version hashes. Each case separates the published result from our interpretation.

\paragraph{E02. Recovery depends on frequency coverage}
\setcounter{evidencecase}{1}\refstepcounter{evidencecase}\label{case:E02}
\textbf{Source and setting:} \citet{N40}: a synthetic dictionary with uniform marginal feature frequencies and variable sample sparsity (Appendix~\ref{app:recovery}).
\textbf{Observation:} at alignment threshold 0.9, TopK recovery is 98.4/24.2/23.4\% for $K=20/30/50$; GBA reaches 100\% with full target-frequency coverage and 3.9\% when coverage excludes the generating frequency.
\textbf{Interpretation:} sparsity allocation and specified frequency coverage change recovery in this known model. The comparison does not measure long-tail linguistic recovery or NLP accuracy.

\paragraph{E10. Direct-space steering can outperform SAE steering}
\setcounter{evidencecase}{9}\refstepcounter{evidencecase}\label{case:E10}
\textbf{Source and setting:} \citet{N67}: CLaS-Bench on Llama-3.1-8B-Instruct across 32 languages (Appendix~\ref{app:steering}).
\textbf{Observation:} residual and SAE DiffMean score 84.5 and 42.3. The reported selected settings use layers 20/25, respectively, with strength 5; the source prose instead gives an SAE peak at strength 15. This discrepancy remains unresolved.
\textbf{Interpretation:} the result favors the selected residual pipeline; it does not isolate dictionary quality from selection and available layers. The SAE edit restores the reconstruction residual, excluding simple reconstruction replacement as an explanation of the gap.

\paragraph{E11. Sparse-probe gains depend on the baseline}
\setcounter{evidencecase}{10}\refstepcounter{evidencecase}\label{case:E11}
\textbf{Source and setting:} \citet{O063}: Gemma-2-9B/JumpReLU GemmaScope, with a Llama-3.1-8B/TopK replication.
\textbf{Observation:} providing attention-pooled dense probes reduces the apparent advantage of max-pooled SAE probes; the source also examines binarization.
\textbf{Interpretation:} baseline construction matters within this protocol. Both compared classification studies examine pooling, so its mere presence cannot reconcile their findings. Appendix~\ref{app:classification} separates the remaining differences in representation, feature selection, supervision, and classifier capacity.

\paragraph{E16. Latent optimization establishes conditional repair}
\setcounter{evidencecase}{15}\refstepcounter{evidencecase}\label{case:E16}
\textbf{Source and setting:} \citet{N55}: frozen models and dictionaries, approximately 255 distractor-vulnerable examples per model (Appendix~\ref{app:rag}).
\textbf{Observation:} optimizing latents toward the correct token recovers answers in this selected population. DCD/MDR comparators instead use oracle document roles. Gemma-2-2B is unambiguous; Llama model naming differs between methods and result labels.
\textbf{Interpretation:} these results establish answer-conditioned repair capacity. They do not evaluate autonomous detection/correction, and post-hoc feature labels do not uniquely identify the natural mechanism.

\paragraph{E17. Sentiment gains vary with the prompt}
\setcounter{evidencecase}{16}\refstepcounter{evidencecase}\label{case:E17}
\textbf{Source and setting:} \citet{N49}: Gemma-2-2B, SST-2, GemmaScope dictionaries, validation-calibrated circuit amplification, and first-output-token classification (Appendix~\ref{app:sentiment}).
\textbf{Observation:} three-run mean F1 gains across T1--T4 are 37.19, 14.99, 6.28, and 2.96 percentage points.
\textbf{Interpretation:} the measured benefit is prompt-dependent; baseline headroom is not isolated as its cause. Zero-shot prompting does not imply unsupervised circuit discovery or strength selection. The extracted means provide no dispersion for significance testing.

\paragraph{E20. Counterexamples refine feature descriptions}
\setcounter{evidencecase}{19}\refstepcounter{evidencecase}\label{case:E20}
\textbf{Source and setting:} \citet{O089}, methods on feature sampling, explanation, activation simulation, and revision: selected features from Gemma-2-9B, Llama-3.1-8B, and GPT-2 Small, assessed with held-out and synthetic tests.
\textbf{Observation:} mismatches between description-predicted and observed activation reveal failures in initial explanations and guide revisions.
\textbf{Interpretation:} this supports a falsifiable description workflow. Description accuracy does not establish a computational role, consistent with the claim distinctions in Table~\ref{tab:rubric} and Appendix~\ref{app:evidence}.

\end{document}